\documentclass{l4dc2025}

\title[FAWAC]{FAWAC: Feasibility Informed Advantage Weighted Regression for Persistent Safety in Offline Reinforcement Learning}
\usepackage{times}
\usepackage{bbm}
\usepackage{amssymb}

\usepackage{graphicx}
\usepackage{subcaption}
\usepackage{wrapfig}
\usepackage{float}
\usepackage{times}
\usepackage{titlesec}

\titlespacing{\section}{0pt}{2pt}{2pt}
\titlespacing{\subsection}{0pt}{2pt}{2pt}

\coltauthor{\Name{Prajwal Koirala} \Email{prajwal@iastate.edu}\\
 \Name{Zhanhong Jiang} \Email{zhjiang@iastate.edu}\\
 \Name{Soumik Sarkar} \Email{soumiks@iastate.edu}\\
 \Name{Cody Fleming} \Email{flemingc@iastate.edu}\\
 \addr Iowa State  University, Ames, Iowa}

\newcommand{\textbbf}[1]{\textbf{\textcolor{blue}{#1}}}
\begin{document}

\maketitle

\begin{abstract}%
Safe offline reinforcement learning aims to learn policies that maximize cumulative rewards while adhering to safety constraints, using only offline data for training. A key challenge is balancing safety and performance, particularly when the policy encounters out-of-distribution (OOD) states and actions, which can lead to safety violations or overly conservative behavior during deployment. To address these challenges, we introduce Feasibility Informed Advantage Weighted Actor-Critic (FAWAC), a method that prioritizes persistent safety in constrained Markov decision processes (CMDPs). FAWAC formulates policy optimization with feasibility conditions derived specifically for offline datasets, enabling safe policy updates in non-parametric policy space, followed by projection into parametric space for constrained actor training. By incorporating a cost-advantage term into Advantage Weighted Regression (AWR), FAWAC ensures that the safety constraints are respected while maximizing performance. Additionally, we propose a strategy to address a more challenging class of problems that involves tempting datasets where trajectories are predominantly high-rewarded but unsafe. Empirical evaluations on standard benchmarks demonstrate that FAWAC achieves strong results, effectively balancing safety and performance in learning policies from the static datasets.
\end{abstract}


\section{Introduction}

Safe offline reinforcement learning (RL) has emerged as a promising approach to addressing the risks associated with \textit{online} RL, where exploration can lead to catastrophic outcomes. In domains like healthcare, autonomous driving, and industrial robotics, any safety violations during training or deployment could result in severe consequences, making it essential to prioritize safety throughout the learning process \citep{garcia2015comprehensive,achiam2017constrained, frye2019parenting, gu2022review}. Safe offline RL seeks to maximize performance while also ensuring that the learned policies adhere to strict safety constraints \citep{xu2022constraints, liu2023datasets}. By balancing performance and stringent safety requirements, this paradigm facilitates robust policy learning for critical applications where online exploration is impractical or too risky.



The safe RL problem is typically studied within the framework of Constrained Markov Decision Processes (CMDPs) \citep{altman1998constrained, altman2021constrained}, with various solution approaches applied to online training. Trust-region methods like Constrained Policy Optimization (CPO) leverage second-order approximations for policy optimization \citep{achiam2017constrained}, while FOCOPS simplifies this by operating in nonparametric policy space and providing an approximate upper bound for constraint violations \citep{zhang2020first}. Lagrangian-based techniques, such as PID Lagrangians, treat constrained RL as a dynamical system to improve stability \citep{stooke2020responsive}. State-wise Lagrange multipliers enhance constraint enforcement by adapting to varying safety requirements across states \citep{ma2021feasible}. \cite{yu2022reachability} further strengthens safety adherence using model-free Hamilton-Jacobi reachability analysis to ensure safe state trajectories.


While these methods have shown promise in online RL, their application to offline RL faces additional challenges, primarily due to distribution shift and often due to the reliance on suboptimal data that may not align with safety constraints \citep{levine2020offline, liu2023datasets}. Safe offline RL requires balancing three competing objectives: maximizing reward, staying close to the behavior policy to mitigate distributional shift, and minimizing cost to adhere to the safety requirements. Recent approaches have attempted to adapt safe online RL methods to the offline setting, but achieving this balance remains challenging, as improvements in one objective often undermine another, with some policies being overly restrictive while others fail to ensure safety \citep{liu2023datasets, liu2023constrained, zheng2024safe}. In this work, we propose Feasibility Informed Advantage Weighted Actor Critic (FAWAC), an actor-critic framework that incorporates feasibility conditions for persistent safety in Constrained Markov Decision Processes (CMDPs). To determine persistent safety for observations in offline datasets, we introduce simplifying assumptions that enable solving the optimization problem in a non-parametric policy space, followed by policy projection into the parametric space. This leads to a supervised learning-style update for training a constrained actor. FAWAC builds on Advantage Weighted Regression (AWR) and its variants in offline and off-policy RL, with a key distinction: we include a cost advantage term in the weight to enforce persistent safety constraints. Evaluations on standard benchmark tasks demonstrate that FAWAC, while simple and intuitive, yields strong results that adhere to the safety criteria.

\begin{figure}
    \centering
    \includegraphics[width=1\linewidth]{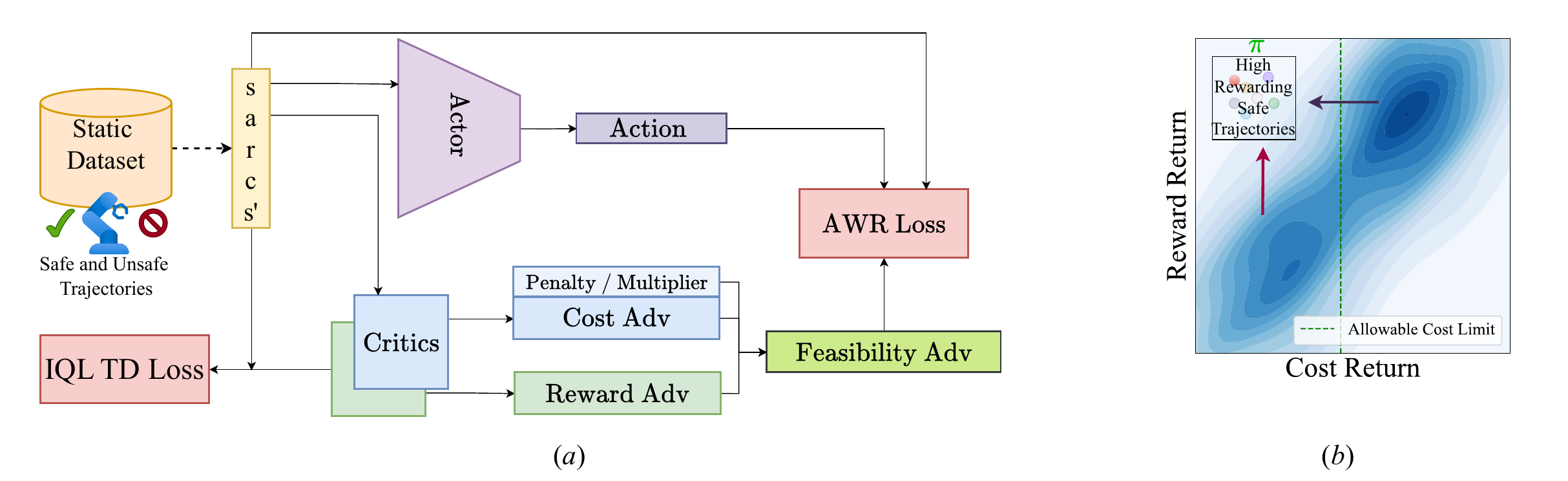}
    \caption{(a) Block diagram illustrating the FAWAC training with a static dataset. Transitions are sampled to train the critic with TD loss and the actor with AWR loss, where the advantage is feasibility-informed, incorporating both cost-advantage and reward standard deviation components.
    (b) Diagram illustrating the distribution of trajectories in the cost-return vs. reward-return space. The goal in safe offline RL is to learn a policy that generates high-performing, safe trajectories.}
    \label{fig:intro-diagram}
\end{figure}


\section{Preliminaries}
\subsection{Safe Offline RL}

Safe Reinforcement Learning (Safe RL) problems are modeled as Constrained Markov Decision Processes (CMDPs), defined by the tuple $\mathcal{M} = (\mathcal{S}, \mathcal{A}, \mathcal{P}, r, c, \gamma, \rho_0)$. Here, $\mathcal{S}$ and $\mathcal{A}$ represent the state and action spaces, $\mathcal{P}: \mathcal{S} \times \mathcal{A} \times \mathcal{S} \to [0, 1]$ is the transition probability function, $r: \mathcal{S} \times \mathcal{A} \to \mathbb{R}$ is the reward function, and $c: \mathcal{S} \times \mathcal{A} \to \mathbb{R}$ is the cost function penalizing unsafe transitions. The discount factor $\gamma \in [0,1)$ and the initial state distribution $\rho_0$ complete the CMDP formulation. 

A trajectory $\tau = {(s_0, a_0, r_0, c_0), (s_1, a_1, r_1, c_1), \dots, (s_T, a_T, r_T, c_T)}$ consists of states, actions, rewards, and costs over time. The discounted cumulative reward is defined as $R(\tau) = \sum_{t=0}^T \gamma^t r(s_t, a_t)$, and the discounted cumulative cost is $C(\tau) = \sum_{t=0}^T \gamma^t c(s_t, a_t)$. The stationary state-action distribution under policy $\pi$ is $d^\pi(s, a) = (1-\gamma) \sum_{t=0}^\infty \gamma^t p(s_t = s, a_t = a)$, where $p(s_t = s, a_t = a)$ is the probability of state $s$ and action $a$ at time $t$, and the stationary state distribution is $d^\pi(s) = \sum_a d^\pi(s, a)$.
The objective in Safe RL is to learn a policy $\pi$ that maximizes the expected return $\mathbb{E}_{\tau \sim \pi}[R(\tau)]$ while ensuring the expected cost $\mathbb{E}{\tau \sim \pi}[C(\tau)]$ satisfies the safety constraint $\mathbb{E}[C(\tau)] \leq \kappa$. 
This is formally stated as:
\begin{equation}\label{eq_1} \max_{\pi} \mathbb{E}[R(\tau)], \quad \text{s.t.} \quad \mathbb{E}[C(\tau)] \leq \kappa. \end{equation}

In offline RL, the learning process is confined to a static dataset, eliminating any active data collection by further interaction with the environment. 
The agent learns the policy from a pre-collected dataset $\mathcal{D}:=(s,a,s',r,c)$, with both safe and unsafe trajectories. The value functions for both reward and cost can be defined in a unified way $V^\pi_h(s_0)=\mathbb{E}_{\tau\sim\pi}[\sum_{t=0}^T\gamma^th_t|s_t=s_0], h\in\{r,c\}$, where $s_0\sim \rho_0$. Similarly, the state-action value $Q_h^\pi(s, a)$ and the advantage $A_h^\pi(s, a) = Q_h^\pi(s, a) - V_h^\pi(s)$ are defined analogously for $h \in \{r, c\}$. Unless explicitly subscripted with $c$, the policy evaluation terms default to reward.
We can now formulate safe offline RL as an optimization problem within the framework of a CMDP:
\begin{equation}\label{eq_2}
    \max_{\pi}V^\pi_r(s),\; s.t.,V^\pi_c(s)\leq \kappa;\;D_{KL}(\pi||\pi_\beta)\leq \delta,
\end{equation}
where $ \pi_\beta $ governs the action distribution in the given dataset, and $ \delta $ is a divergence tolerance parameter. The function $ D_{KL}(\pi||\pi_\beta) $ measures the divergence between the learned policy $ \pi $ and the behavior policy $ \pi_\beta $, typically using metrics such as Kullback-Leibler (KL) divergence or other statistical distance measures. These constraints ensure that the learned policy $ \pi $ remains within the safety limits defined by the cost threshold $ \kappa $ and does not deviate significantly from the behavior policy $ \pi_\beta $, thereby mitigating the risks associated with out-of-distribution actions. 

\subsection{Implicit Q-Learning}
Implicit Q-Learning (IQL) uses an asymmetric loss function in a SARSA-style policy evaluation framework to train a state-value network $V_\phi$, mitigating the impact of out-of-distribution actions during critic training \citep{kostrikov2021iql}. The critic networks are trained with the TD loss:
\begin{align}
\mathcal{L}_Q(\mu) &= \mathbb{E}_{(s,a,s') \sim \mathcal{D}} \left[ \left( r + \gamma V_\phi(s') - Q_\mu(s, a) \right)^2 \right] \label{eq:q_loss_reward} \\
\mathcal{L}_V(\phi) &= \mathbb{E}_{(s,a) \sim \mathcal{D}} \left[ L_\tau^2 \left( Q_{\mu}(s, a) - V_\phi(s) \right) \right], \label{eq:v_loss_reward}
\end{align}
where $\mu$ and $\phi$ are model parameters.
The value network \( V_\phi \) is trained via an expectile regression objective, with the asymmetric squared loss \( L_\tau^2(u) = |\tau - 1(u < 0)| u^2 \), where \( \tau \in (0.5, 1.0) \). The learned advantage is given by $A(s, a) = Q_\mu(s, a) - V_\phi(s)$ which is commonly used for policy extraction through Advantage-Weighted Regression (AWR) \citep{peng2019advantage}. For cost-critic training, we extend IQL to learn cost-value functions. The cost-specific $Q$ and value networks, \( Q^c_\psi \) and \( V^c_\eta \), parameterized by $\eta$ and $\psi$, are trained using:
\begin{align}
\mathcal{L}_Q^c(\psi) &= \mathbb{E}_{(s,a,s') \sim \mathcal{D}} \left[ \left( c + \gamma V^c_\eta(s') - Q^c_\psi(s, a) \right)^2 \right] \label{eq:q_loss_cost} \\
\mathcal{L}_V^c(\eta) &= \mathbb{E}_{(s,a) \sim \mathcal{D}} \left[ L_\tau^2 \left( V^c_\eta(s) - Q^c_\psi(s, a) \right) \right] \label{eq:v_loss_cost}
\end{align}

\subsection{Safety and Feasibility}
While $c(s,a)$ gives the immediate constraint violation cost, we define $c(s,a)=0$ as temporary safety. For persistent safety under budget $\kappa$, we define the feasible region for a policy $\pi$ as follows:
\begin{equation}
\mathcal{S}_f^\pi = \{s \in \mathcal{S} \mid V_c^\pi(s) \leq \kappa\}
\end{equation}
The largest feasible set $\mathcal{S}_f$ is a subset of $\mathcal{S}$ from which there exists at least one policy for which the feasible region is a non-empty set, i.e., there exists a policy that can keep the persistent safety condition $V_c^\pi(s) \leq \kappa$:
\begin{equation}
\mathcal{S}_f = \{s \in \mathcal{S} \mid \exists \pi, V_c^\pi(s) \leq \kappa\} \label{feasible region equation}
\end{equation}
Denote the set of all the states in the dataset $\mathcal{D}$ as $\mathcal{S}_0$ and the largest feasible set within these states as $\mathcal{S}_{f0} = \mathcal{S}_0 \cap \mathcal{S}_f$. Then, conversely, we can define a feasible policy set as:
\begin{equation}
\Pi_f = \left\{ \pi \mid V_c^\pi(s) \leq \kappa, \forall s \in \mathcal{S}_{f0}, \, \text{and} \, D_{\text{KL}}(\pi \parallel \pi_\beta) \leq \delta \right\}
\end{equation}
Formally, we formulate the following optimization  problem to solve the CMDP described in the previous section:
\begin{align}
\max_{\pi} & \quad \mathbb{E}_{s \sim \mathcal{D}}\left[V_r^\pi(s) \cdot \mathbbm{1}_{s \in \mathcal{S}_f} - V_c^\pi(s) \cdot \mathbbm{1}_{s \notin \mathcal{S}_f}\right] \label{feasible-infeasible-optim} \\
\text{subject to} & \quad V_c^\pi(s) \leq \kappa, \quad \forall s \in \mathcal{S}_{f0} \\
& \quad D_{\text{KL}}(\pi \parallel \pi_\beta) \leq \delta \label{feasible-infeasible-optim-ineq-2}
\end{align}
While we can determine the feasible region for policy $\pi$ as given by equation \ref{feasible region equation}, the largest feasible region is not always known. This makes directly solving the above optimization problem extremely challenging. To mitigate this issue, we resort to a method that approximates \textit{feasibility advantage} and helps us frame this as an advantage weighted regression (AWR) problem.

\section{Methodology and Analysis}
In this section, we present the technical details about our proposed algorithms and the main formal results. Due to the space limitation, we defer the relevant analysis of these results to the Appendix in section \ref{appendix_section}.
In what follows, we start with the feasibility advantage weighted regression.
\subsection{Feasibility Advantage Weighted Regression}
To simplify/approximate the solution to the optimization problem in equations \ref{feasible-infeasible-optim}-\ref{feasible-infeasible-optim-ineq-2}, we introduce two assumptions: $\mathcal{S}_{f0} = \mathcal{S}_{0}$ and an initial policy $\pi_0 \in \Pi_f$. These two assumptions are only taken to ease the problem formulation and the relevant analysis. In practice, the first assumption may not strictly hold always as the offline dataset $\mathcal{D}$ can still comprise data samples produced by infeasible or unsafe policies.
Then, the sequential policy update rule can be written as the following optimization problem from Eqs.~\ref{feasible-infeasible-optim}-\ref{feasible-infeasible-optim-ineq-2}:
\begin{align}
\pi_{k+1} = \arg \max_{\pi} & \quad \mathbb{E}_{ a \sim \pi} \left[ A^{\pi^k}(s, a) \right] \label{achiam-adv-optim-problem}\\
\text{s.t.} & \quad V^{\pi_k}_{c}(s) + \frac{1}{1-\gamma} \mathbb{E}_{a \sim \pi} \left[ A_{c}^{\pi^k}(s, a) \right] \leq \kappa, \label{eq_14}\\
& \quad D_{KL}(\pi || \pi_\beta) \leq \delta, \label{eq_15}\\
& \quad \int_a \pi(a|s) da = 1, \label{achiam-adv-optim-problem-probab-constraint}
\end{align}
where $\pi^k$ is the currently accessible policy. With abuse of terminology, we use $\delta$ as the upper bound for the behavior regularization constraint. Please see Appendix for more details regarding this approximation, which is a non-trivial effort.
Similar formulations have been used to derive on-policy RL algorithms with a constraint on trust region $D_{KL}(\pi || \pi^k)$ \citep{achiam2017constrained, zhang2020first}. However, different from these existing works, they have primarily focused on online settings. 
Instead, in this study, we make an extension from the online to offline setting and strategically use a behavior regularization constraint to address distribution shift in offline RL. Next, we solve the above approximate optimization problem by first searching for the \textit{non-parameterized} optimal policy $\pi^*$, given the current policy $\pi^k$. It should be noted that $\pi^k$ is equivalent to $\pi_{\theta_k}$ when a parameterized model $\theta$ is used to model the policy. To characterize the analysis for the optimization, we present the following lemma to show the problem in Eqs.~\ref{achiam-adv-optim-problem} to~\ref{achiam-adv-optim-problem-probab-constraint} is a \textit{convex} optimization problem.
\begin{lemma}\label{lemma_1}
    The optimization problem induced by Eqs.~\ref{achiam-adv-optim-problem} to~\ref{achiam-adv-optim-problem-probab-constraint} is a convex optimization problem.
\end{lemma}
Lemma~\ref{lemma_1} ensures the solvability of the problem either theoretically or practically, which assists in the derivation of the optimal solution $\pi^*$. 
Let $\bar{\kappa} = (\kappa - V^{\pi^k}_{c}(s)) (1-\gamma)$. As the optimization problem is constrained, its Lagrangian can be written as:
\begin{align}
\mathcal{L}(\pi, \nu, \lambda, \alpha) = &\mathbb{E}_{a \sim \pi(\cdot|s)} \left[ A^{\pi^k}(s, a) \right] 
+ \nu \left( \bar{\kappa} - \mathbb{E}_{a \sim \pi} \left[ A_{c}^{\pi^k}(s, a) \right] \right) + \cdots
 \nonumber \\ 
& + \lambda \left( \delta - D_{KL}(\pi(\cdot|s) \| \pi_\beta(\cdot|s)) \right) + \alpha \left( 1 - \int_a \pi(a|s) \, da \right),
\end{align}
where $\nu, \lambda, \alpha$ are the multipliers. Hence, we can arrive at a main result in the following.
\begin{theorem}\label{theorem_2}
    Let $\bar{\kappa} = (\kappa - V^{\pi^k}_{c}(s)) (1-\gamma)$. If $\pi^k$ is a feasible solution, the optimal policy for the optimization problem in Eqs.~\ref{achiam-adv-optim-problem} to~\ref{achiam-adv-optim-problem-probab-constraint} is in the following form
    \begin{equation}\label{eq_18}
        \pi^*(a|s) = \frac{1}{Z(s)} \pi_\beta(a|s) \exp\left(\frac{1}{\lambda} \left( A^{\pi^k}(s, a) - \nu A_c^{\pi^k}(s, a) \right) \right),
    \end{equation}
where $Z(s)$ is a partition function to ensure $\pi^*(a|s)$ is a valid probability distribution function, which is a function w.r.t. $\lambda$ and $\alpha$.
\end{theorem}
Despite the fact that this form of the solution has been derived in related works on either off-policy~\citep{peng2019advantage} or offline~\citep{nair2020awac} RL and safe RL~\citep{zhang2020first}, we are not aware of any reported results on the offline safe RL, which is a more challenging problem. The term $A^{\pi^k}(s, a) - \nu A_c^{\pi^k}(s, a)$ represents a \textit{feasibility-informed} advantage, as it incorporates safety considerations into the advantage estimation, ensuring optimization over a feasible policy set. $\pi^*$ intuitively imposes high probability mass to the space of state-action pairs in $\mathcal{D}$ that yields high return, which is also balanced by a penalized cost advantage $A^{\pi^k}_c(s,a)$. In this context, we have only one constraint, while the problem formulation is fairly flexible to accommodate for multiple constraints that may or may not be related to cost, by inaugurating multiple Lagrangian multipliers. The constrained optimization problem in Eqs.~\ref{achiam-adv-optim-problem} to~\ref{achiam-adv-optim-problem-probab-constraint} requires the safety constraint induced by the cost state value $V^{\pi_k}_c(s)$ and the cost advantage $A^{\pi^k}_c$. Therefore, the investigation to the constraint violation is central to the policy learning. Luckily, the optimal policy $\pi^*$ offers a desirable property that the following upper bound holds for the worst-case guarantee given feasible policy $\pi^{k}\in\Pi_f$.
\begin{proposition}\label{proposition_3}
    Suppose that $\pi^*$ is produced by Eq.~\ref{eq_18}. Then for any given feasible policy $\pi^{k}\in\Pi_f$, it holds true that
    $
        \mathbb{E}_{s\in\mathcal{S}_0}[V^{\pi^*}_c(s)]\leq \kappa +\frac{2\sqrt{2\delta}\gamma\epsilon^{\pi^*}_c}{(1-\gamma)^2}
   $, where $\epsilon^{\pi^*}_c=\textnormal{max}_{s\in\mathcal{S}_0}|\mathbb{E}_{a\sim\pi^*}[A^{\pi^{k}}_c(s,a)]|$.
\end{proposition}
In view of the constraint violation bound, we notice that the second term on the right hand side is larger than that in~\citep{zhang2020first}, which is attributed to the data distribution shift between the unknown behavior policy $\pi_\beta$ and the learned policy $\pi$.
We are now ready to derive the parametric policy $\pi_\theta$ by projecting $\pi^*$ back to the parameterized policy update and solving another minimization problem between $\pi^*$ and $\pi_\theta$. 

\paragraph{Parametric policy $\pi_\theta$ update.}
When the policies are parametrized, the optimal policy $\pi^*$ can be projected onto the manifold of parametrized policies by minimizing the KL divergence between a policy $\pi_\theta$ with parameters $\theta$ and the optimal solution $\pi^*$. Immediately, $\pi^{\theta_k}:=\pi^k$ such that:
\begin{equation}\label{para_policy_update}
\begin{split}
\theta_{k+1}=&\arg \min_{\theta} \, \mathbb{E}_{s \sim d^{\pi_\beta}(s)} \left[ D_{KL}(\pi^*(\cdot|s) \| \pi_\theta(\cdot|s)) \right] \\
=&\arg \min_{\theta} \, \mathbb{E}_{s \sim d^{\pi_\beta}(s)} \left[ D_{KL}\left(\frac{1}{Z(s)} \pi_\beta(a|s) \exp\left(\frac{1}{\lambda} \left( A^{\pi^{\theta_k}}(s, a) - \nu A_c^{\pi^{\theta_k}}(s, a) \right) \right) \| \pi_\theta(\cdot|s)\right) \right] \\
\approx&\arg \min_{\theta} \, \mathbb{E}_{s \sim d^{\pi_\beta}(s)} \mathbb{E}_{a \sim \pi_\beta(a|s)} \left[ - \exp\left(\frac{1}{\lambda} \left( A^{\pi^{\theta_k}}(s, a) - \nu A_c^{\pi^{\theta_k}}(s, a) \right) \right) \log \pi_\theta(\cdot|s) \right]
\end{split}
\end{equation}
The last equality follows similarly from~\citep{peng2019advantage}. In the Lagrangian $\mathcal{L}(\pi, \nu, \lambda, \alpha)$, $\alpha$ is the multiplier for $\int_a \pi(a|s) da = 1$, which only affects $Z(s)$, such that during training, we can pre-define this parameter, instead of continuously adapting it. $\lambda$ is another key multiplier that controls the feasibility-informed advantage in $\pi^*$. Intuitively, when $\lambda \to 0$, $\pi^*$ gradually becomes a vanilla greedy policy; on the contrary, it is more exploratory when $\lambda$ increases. Essentially, $\lambda$ is similar to the \textit{temperature} term utilized in the maximum entropy-based RL~\citep{ziebart2008maximum,aghasadeghi2011maximum}, which has empirically been shown to yield satisfactory results when it is a fixed value during learning~\citep{schulman2017equivalence,haarnoja2018soft}. Thus, 
we pre-define $\lambda$ so as to reduce the complexity of solving the optimization problem for $\theta$. The most critical multiplier is $\nu$ that penalizes the cost advantage in the non-parameterized optimal policy and has an impact on the ultimately optimal parameter $\theta^*$. It requires continuous updates to provide the assurance of cost constraint. Thereby, we discuss two distinct approaches for handling the multiplier $\nu$.
\paragraph{Multiplier $\nu$ update: Direct optimization.}
The first approach to update the multiplier $\nu$ is to conduct the direct optimization. We tactically employ the primal-dual gradient method~\citep{xu2020primal} that has widely been used to solve constrained optimization problems~\citep{gill2012primal}. Our first step is to show that $\mathcal{L}$ has strong duality. In~\citep{zhang2020first}, their problem formulation enables the strong duality for the Lagrangian, as their analysis has centered on the online setting such that the distance constraint in Eq.~\ref{eq_15} simply degenerates to $D_{KL}( \pi^{\theta_k}||\pi^{\theta_k})= 0$, which helps determine $\pi^{\theta_k}$ as an \textit{interior} point with conditions that $\delta>0$ and that the optimization problem itself is convex. However, the setting in this study is offline such that the distribution shift between the unknown behavior policy $\pi_\beta$ and the learned policy $\pi$ causes difficulty for $\pi^{\theta_k}$ to be interior, even if it is feasible. To alleviate this issue, we assume throughout the analysis that $D_{KL}( \pi^{\theta_k}||\pi_\beta)< \delta$, $\pi^{\theta_k}\in\Pi_f$. This assumption suggests that there exist some points in the set $\Pi_f$ that are always interior yet feasible. Additionally, since $\pi^{\theta_k}$ satisfies Eq.~\ref{eq_14} and Lemma~\ref{lemma_1} holds, the Slater's constraint qualification condition~\citep{dempe1993directional} holds, which leads to the \textit{strong duality}. With this in hand, the optimization problem in Eqs.~\ref{achiam-adv-optim-problem} to~\ref{achiam-adv-optim-problem-probab-constraint} can accordingly be solved via its dual problem.
\begin{equation}\label{eq_20}
    \underset{\pi\in\Pi_f}{\text{max}}\;\underset{\nu\geq 0}{\text{min}}\;\mathcal{L}(\pi,\nu,\lambda,\alpha) = \underset{\nu\geq 0}{\text{min}}\;\underset{\pi\in\Pi_f}{\text{max}}\;\mathcal{L}(\pi,\nu,\lambda,\alpha)
\end{equation}
Therefore, if $\pi^*$ and $\nu^*$ are optimal for Eq.~\ref{eq_20}, $\pi^*$ is also optimal for the optimization problem in Eqs.~\ref{achiam-adv-optim-problem} to~\ref{achiam-adv-optim-problem-probab-constraint}~\citep{boyd2004convex}. Denoting by $\nu^*$ the optimal solution of $\nu$,
we have: 
\begin{align}
\nu^* = \arg\min_{\nu\geq0} \left[ \nu \left( \bar{\kappa} - \mathbb{E}_{a \sim \pi^{\theta_{k+1}}} \left[ A_{c}^{\pi^{\theta_k}}(s, a) \right] \right) \right] \label{nu_star_dual_optimization}
\end{align}
Unfortunately, the expectation of $A_{c}^{\pi^{\theta_k}}(s, a)$ cannot be evaluated as $\pi^{\theta_{k+1}}$ is inaccessible, as the update to $\theta$ is after the update to $\nu$. However, $\pi^{\theta_k}$ and $\pi^{\theta_{k+1}}$ are  ``close" as their KL divergence with the behavior policy $\pi_\beta$ is constrained, we thus assume that:
\begin{align}
\mathbb{E}_{a \sim \pi_{\theta_{k+1}}} \left[ A_{c}^{\pi^{\theta_k}}(s, a) \right] \approx \mathbb{E}_{a \sim \pi^{\theta_{k}}} \left[ A_{c}^{\pi^{\theta_k}}(s, a) \right] = 0 \label{approx_cost_adv}
\end{align}
With this approximation, the relationship in \ref{nu_star_dual_optimization} simplifies to:
\begin{align}\label{nu_star_dual_optimization_simplified}
\nu^* = \arg\min_{\nu\geq0} \left( \nu \bar{\kappa} \right) = \arg\min_{\nu\geq0} \left( \nu (\kappa - V^{\pi^k}_{c}(s)) \right) = \arg\max_{\nu\geq0} \left( \nu (V^{\pi^k}_{c}(s) - \kappa) \right)
\end{align}
To practically update $\nu$, we can use the projected gradient descent to ensure that the updated $\nu$ is always within a reasonable range. Intuitively, Eq.~\ref{eq_18} implies that $\nu$ penalizes $\pi^*$ when it samples state-action pairs with higher costs. This complies with the implication from Eq.~\ref{nu_star_dual_optimization_simplified} that $V^{\pi^k}_{c}(s) > \kappa$ results in a larger $\nu$, while $V^{\pi^k}_{c}(s) < \kappa$ provokes a smaller one. 

Recent works \citep{ma2021feasible, yu2022reachability} have proposed a more general way of imposing state-wise constraints using a parametrized approximation to the multiplier as $\nu_\zeta: \mathcal{S} \rightarrow \mathbb{R}_{\geq0}$, where $\zeta$ are learnable parameters for the multiplier $\nu(s)$. A parametrized multiplier provides a localized mechanism for indicating and handling state-wise feasibility. Specifically, \( \nu_\zeta(s) = 0 \) deactivates the constraint for a given state, while a finite value of \( \nu_\zeta(s) \) signifies that the constraint is active and the state lies on the boundary of the feasible region. Conversely, an infinite value indicates that the state is infeasible, thereby directing the optimization towards cost minimization for that state. In practice, however, we use a threshold \( \nu_{\text{max}} \) to bound the values of \( \nu_\zeta(s) \) to effectively handle the infinite values. We now have the following optimization problem for \( \zeta \):
\begin{align}
\zeta^* = \arg \min_{\zeta} \nu_\zeta(s) (\kappa - V^{\pi^{\theta_k}}_{c}(s))
\end{align}


\paragraph{Finite penalty on cost constraint.}
In the above formulation, we have resorted to the primal-dual gradient method to solve the optimization problem in Eqs.~\ref{achiam-adv-optim-problem} to~\ref{achiam-adv-optim-problem-probab-constraint}. Though this approach is computationally compelling, it typically requires the strong duality to hold, which may not necessarily be true in diverse scenarios. Also, we need to implement optimization to the multiplier. If there are numerous constraints, the optimization can be extremely complex and not practically tractable. Alternatively, the multiplier optimization can be bypassed by using a finite penalty factor $\hat{\nu}$ instead \citep{zhang2022penalized, zhang2023evaluating}. Specifically, we reformulate the multiplier \(\nu(s)\) to take values of either \(0\) or \(\hat{\nu}\), depending on the feasibility condition, thereby simplifying the optimization process.
\begin{align*}
\mathcal{L}'(\pi, \hat{\nu}, \lambda, \alpha) = \mathbb{E}_{a \sim \pi(\cdot|s)} \left[ A^{\pi^k}(s, a) \right] 
+ \hat{\nu} \min \left\{ \bar{\kappa} - \mathbb{E}_{a \sim \pi} \left[ A_{c}^{\pi^k}(s, a) \right], 0 \right\} \\
+ \lambda \left( \delta - D_{KL}(\pi(\cdot|s) \| \pi_\beta(\cdot|s)) \right)  
+ \alpha \left( 1 - \int_a \pi(a|s) \, da \right).
\end{align*}
With this, the corresponding parametric policy update rule can be derived as:
\begin{align}\label{eq_26}
\theta_{k+1}=&\arg \min_{\theta} \, \mathbb{E}_{s \sim d^{\pi_\beta}(s)} \mathbb{E}_{a \sim \pi_\beta(a|s)} \left[ - \exp\left(\frac{1}{\lambda} \left( A^{\pi^{\theta_k}}(s, a) - \hat\nu A_c^{\pi^{\theta_k}}(s, a) \mathcal{I}_c \right) \right) \log \pi_\theta(\cdot|s) \right],
\end{align}
    where $\mathcal{I}_c = \mathbbm{1}_{\bar{\kappa} - \mathbb{E}_{a \sim \pi} \left[ A_{c}^{\pi^{\theta_k}}(s, a) \right] \leq 0} $ is an indicator function. From Eq.~\ref{approx_cost_adv}, we can approximate $\mathbb{E}_{a \sim \pi_{\theta_{k+1}}} \left[ A_{c}^{\pi_{\theta_k}}(s, a) \right] \approx0$
such that $
    \mathcal{I}_c \approx \mathbbm{1}_{\bar{\kappa} \leq 0} = \mathbbm{1}_{\kappa - V^{\pi_{\theta_k}}_{c}(s) \leq 0}
$.
Comparing Eq.~\ref{eq_26} to Eq.~\ref{para_policy_update} infers that the costs advantage incurred due to the constraint violation is simply controlled by an indicator function $\mathbbm{1}_{\kappa - V^{\pi_{\theta_k}}_{c}(s) \leq 0}$, which simplifies the update for $\theta$, without depending on the search for $\nu$. However, this also prompts the selection of the penalty term $\hat{\nu}$ that possibly requires some manual tuning efforts. Surprisingly, our empirical evidence will show that this simple finite penalty method outperforms the direct optimization.
\subsection{Feasibility Informed Advantage Weighted Actor-Critic Framework}
The Feasibility Informed Advantage Weighted Actor-Critic (FAWAC) framework integrates constrained actor updates with a value-based critic to enforce persistent safety in offline reinforcement learning. The actor update minimizes the KL divergence between the parameterized policy $\pi_\theta$ and the optimal non-parametric policy $\pi^*$, using a feasibility-informed weighted behavior cloning. This ensures that the learned policy adheres to constraints on safety—keeping the cost value below a specified threshold—and on behavior regularization, by maintaining a bounded KL divergence with the behavior policy, while maximizing performance. The critic, based on Implicit Q-Learning (IQL), estimates the reward and cost advantages necessary for the actor update.

FAWAC introduces two methods for handling the feasibility constraint in the policy extraction: (1) \textbf{FAWAC-M}, which employs a statewise multiplier $\nu_\zeta(s)$ learned through primal-dual optimization, dynamically enforcing constraints; and (2) \textbf{FAWAC-P}, which approximates the multiplier using a fixed penalty factor $\hat{\nu}$, directly penalizing safety violations. Policy update for FAWAC-M is based on equation \ref{para_policy_update} and that for FAWAC-P is based on equation \ref{eq_26}.


\subsection{Tempting Dataset, and FAWAC-T}

In safe offline RL, a `tempting' dataset predominantly comprises high-reward trajectories that violate safety constraints \citep{yao2024oasis}. For $s \sim \mathcal{D}$, there is a high chance that $s \notin \mathcal{S}_f^{\pi_\beta}$ due to the nature of the dataset, making the assumption $\mathcal{S}_{f0} = \mathcal{S}_{0}$ unrealistic as policy search space is restricted to be near the behavior policy in offline RL. Instead, assuming that such $s \notin \mathcal{S}_f$, the policy update rule in the parametrized space specific to the tempting dataset can be derived as:
\begin{align*}
\theta_{k+1}=& \arg \min_{\theta} \, \mathbb{E}_{s \sim d^{\pi_\beta}(s)} \mathbb{E}_{a \sim \pi_\beta(a|s)} \left[ - \exp\left(\frac{-1}{\lambda}  A_c^{\pi^{\theta_k}}(s, a) \right) \log \pi_\theta(\cdot|s) \right]
\end{align*}
This update does not explicitly optimize for reward improvement, but given the dataset's nature, it is reasonable to assume $V_r^{\pi_\beta} \geq V_r^{\pi^*}$ due to the tempting nature of the dataset. We refer to this method as \textbf{FAWAC-T}, which is evaluated on tempting datasets against recent baselines. The temperature hyperparameter $\lambda$ regulates the trade-off between adhering to high-rewarding (potentially unsafe) behavior and minimizing the expected cost return.


\section{Experiments}

\subsection{Evaluation Metric and Datasets}

Our evaluation utilizes the DSRL benchmark \citep{liu2023datasets}, which assesses performance based on normalized reward return ($R$) and normalized cost return ($C$) across robot control task sets like the Bullet Safety Gym \citep{gronauer2022bullet}. These tasks, implemented in PyBullet \citep{coumans2016pybullet}, include Run and Circle with robot variants like Ball, Car, and Drone. Normalized reward and cost for an episode are computed as $R = \frac{R_\pi - R_{\text{min}}}{R_{\text{max}} - R_{\text{min}}}$ and $C = \frac{C_\pi}{\kappa'}$, where $R_\pi$ and $C_\pi$ are the accumulated rewards and costs, $R_{\text{min}}$ and $R_{\text{max}}$ are task-specific constants, and $\kappa'$ is the target cost threshold. Three thresholds ($\kappa' \in {10, 20, 40}$) and three random seeds are used to report the average normalized reward and cost. 

For the tempting datasets, we follow the convention of \cite{yao2024oasis}. Specifically, the full dataset from DSRL is modified by randomly filtering out 90\% of the trajectories within the cost interval $[0,30]$ and setting the target cost threshold ($\kappa'$) to $20$.


\subsection{Results}
\begin{figure}[hbt!]
\centering 
\subfigure[Ball Circle]{\label{fig:TrainingCurvesb}\includegraphics[width=75mm]{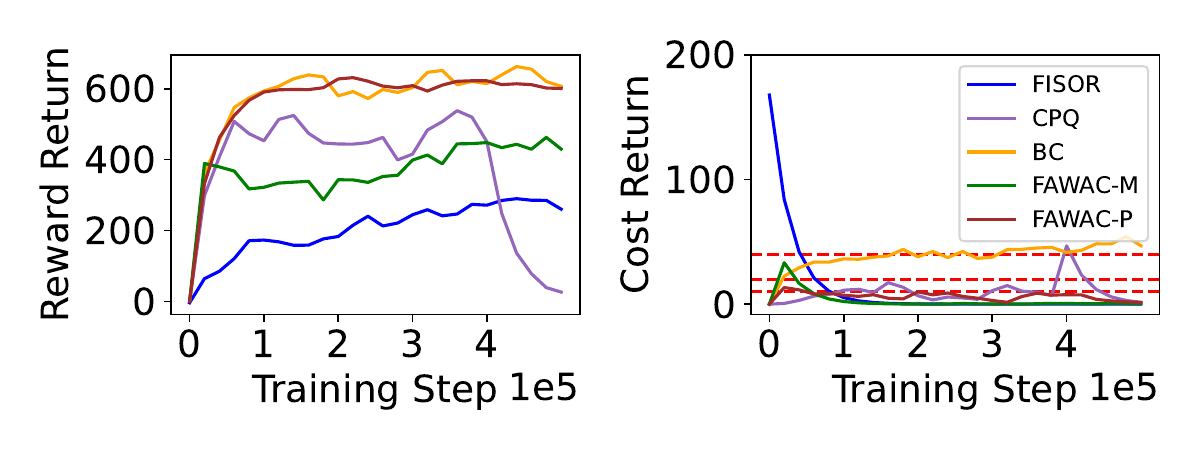}}
\subfigure[Car Circle]{\label{fig:TrainingCurvesa}\includegraphics[width=75mm]{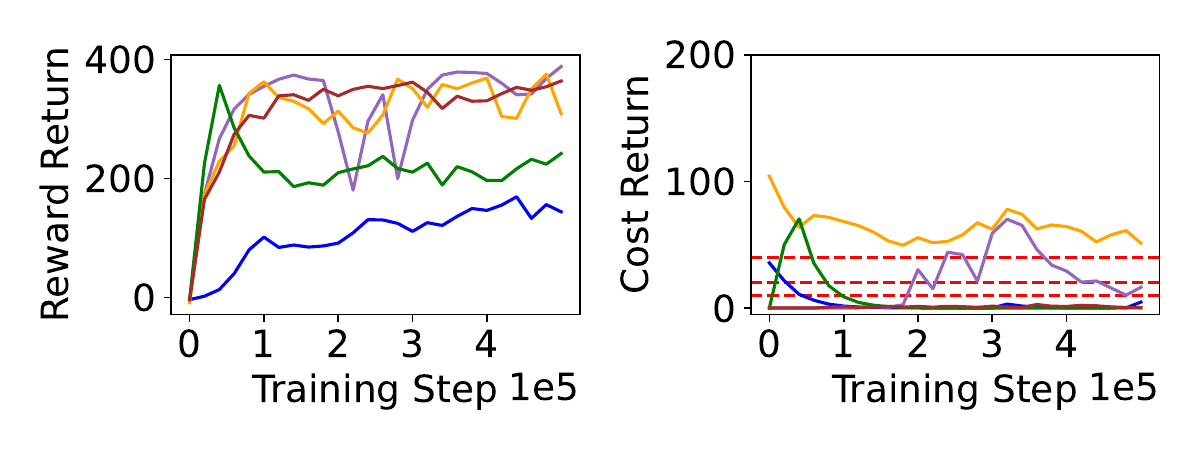}}
\vspace{-7pt}
\caption{Training curves depicting reward and cost returns for our method and baselines.}
\label{fig:TrainingCurves}
\end{figure}

In our evaluation, a normalized cost threshold of 1 is used. Safe agents ($C < 1$) are highlighted in bold, with the high-performing safe agents marked in bold blue. Each method is evaluated across three random seeds, with 20 evaluations per seed, ensuring robust statistical significance. Table \ref{tab:performance_large_table} reports performance on the full dataset, while Table \ref{tab:performance_large_table_tempting} focuses on tempting datasets. FAWAC methods consistently outperform baselines in both settings. Figure \ref{fig:TrainingCurves} shows training curves for FAWAC-M, FAWAC-P, and baselines on Ball Circle and Car Circle tasks, demonstrating the trade-off between safety compliance and reward optimization.

\begin{table}[h]
    \centering
    \caption{Comparison of our methods and baselines on full datasets.} \label{tab:performance_large_table}
    \vspace{-10pt}
    \resizebox{1.0\textwidth}{!}{
    \begin{tabular}{|c|cc|cc|cc|cc|cc|cc||cc|}
        \hline
        \textbf{Task} & \multicolumn{2}{c}{\textbf{BallRun}} & \multicolumn{2}{c}{\textbf{CarRun}} & \multicolumn{2}{c}{\textbf{DroneRun}} & \multicolumn{2}{c}{\textbf{BallCircle}} & \multicolumn{2}{c}{\textbf{CarCircle}} & \multicolumn{2}{c||}{\textbf{DroneCircle}} & \multicolumn{2}{c|}{\textbf{Average}} \\
        \hline
        Method & reward $\uparrow$ & cost $\downarrow$ & reward $\uparrow$ & cost $\downarrow$ & reward $\uparrow$ & cost $\downarrow$ & reward $\uparrow$ & cost $\downarrow$ & reward $\uparrow$ & cost $\downarrow$ & reward $\uparrow$ & cost $\downarrow$ & reward $\uparrow$ & cost $\downarrow$ \\
        \hline
        BC      & 0.60 & 5.08 & \textbbf{0.97} & \textbbf{0.33} & 0.24 & 2.13 & 0.74 & 4.71 & 0.58 & 3.74 & 0.72 & 3.03 & 0.66 & 3.17 \\
        CDT     & 0.39 & 1.16 & \textbbf{0.99} & \textbbf{0.65} & \textbbf{0.63} & \textbbf{0.79} & 0.77 & 1.07 & \textbbf{0.75} & \textbbf{0.95} & \textbbf{0.60} & \textbbf{0.98} & 0.69 & 0.93 \\
        CPQ     & 0.22 & 1.27 & 0.95 & 1.79 & 0.33 & 3.52 & \textbf{0.64} & \textbf{0.76} & \textbf{0.71} & \textbf{0.33} & -0.22 & 1.28 & 0.49 & 1.49 \\
        FISOR   & \textbf{0.18} & \textbf{0.00} & \textbf{0.73} & \textbf{0.04} & \textbf{0.30} & \textbf{0.16} & \textbf{0.34} & \textbf{0.00} & \textbf{0.40} & \textbf{0.03} & \textbf{0.48} & \textbf{0.00} & \textbf{0.41} & \textbf{0.04} \\
        COptiDICE & 0.59 & 3.52 & \textbf{0.89} & \textbf{0.00} & 0.67 & 4.15 & 0.70 & 2.61 & 0.49 & 3.14 & 0.26 & 1.02 & 0.60 & 2.41 \\
        \hline
        FAWAC-M & \textbbf{0.29} & \textbbf{0.76}  & \textbbf{0.99} & \textbbf{0.42} & \textbf{0.47} & \textbf{0.00}  & \textbf{0.54} & \textbf{0.11} &  \textbf{0.43} & \textbf{0.00} & \textbf{0.29} & \textbf{0.05}  & \textbf{0.50} & \textbf{0.22} \\
        FAWAC-P & \textbf{0.21} & \textbf{0.06} & \textbbf{0.98} & \textbbf{0.02} & \textbf{0.50} & \textbf{0.00} & \textbbf{0.71} & \textbbf{0.78} & \textbf{0.68} & \textbf{0.12} & \textbf{0.44} & \textbf{0.02} & \textbbf{0.59} & \textbbf{0.17} \\
        \hline
    \end{tabular}
    }
\end{table}
\vspace{-10pt}
\begin{table}[h!]
    \centering
    \caption{Comparison of our method and baselines on tempting datasets.} \label{tab:performance_large_table_tempting}
    \vspace{-10pt}
    \resizebox{1.0\textwidth}{!}{
    \begin{tabular}{|c|cc|cc|cc|cc|cc|cc||cc|}
        \hline
        \textbf{Task} & \multicolumn{2}{c}{\textbf{BallRun}} & \multicolumn{2}{c}{\textbf{CarRun}} & \multicolumn{2}{c}{\textbf{DroneRun}} & \multicolumn{2}{c}{\textbf{BallCircle}} & \multicolumn{2}{c}{\textbf{CarCircle}} & \multicolumn{2}{c||}{\textbf{DroneCircle}} & \multicolumn{2}{c|}{\textbf{Average}} \\
        \hline
        Method & reward $\uparrow$ & cost $\downarrow$ & reward $\uparrow$ & cost $\downarrow$ & reward $\uparrow$ & cost $\downarrow$ & reward $\uparrow$ & cost $\downarrow$ & reward $\uparrow$ & cost $\downarrow$ & reward $\uparrow$ & cost $\downarrow$ & reward $\uparrow$ & cost $\downarrow$ \\
        \hline
        BC      & 0.55 & 2.04 & 0.94 & 1.50 & 0.62 & 3.48 & 0.73 & 2.53 & 0.59 & 3.39 & 0.82 & 3.29 & 0.71 & 2.71 \\
        CDT     & 0.35 & 1.56 & \textbbf{0.96} & \textbbf{0.67} & 0.84 & 7.56 & 0.73 & 1.36 & 0.71 & 2.39 & 0.17 & 1.08 & 0.63 & 2.44 \\
        CPQ     & 0.25 & 1.34 & 0.63 & 1.43 & 0.13 & 2.29 & \textbf{0.39} & \textbf{0.73} & \textbf{0.64} & \textbf{0.12} & 0.01 & 3.16 & 0.34 & 1.51 \\
        FISOR   & \textbf{0.17} & \textbf{0.04} & \textbf{0.85} & \textbf{0.15} & 0.44 & 2.52 & \textbf{0.28} & \textbf{0.00} & \textbf{0.24} & \textbf{0.15} & \textbf{0.49} & \textbf{0.02} & \textbf{0.41} & \textbf{0.48} \\
        COptiDICE & 0.63 & 3.13 & \textbf{0.90} & \textbf{0.28} & 0.71 & 3.87 & 0.73 & 2.83 & 0.52 & 3.56 & \textbf{0.35} & \textbf{0.12} & 0.64 & 2.30 \\
        OASIS & \textbbf{0.28} & \textbbf{0.79} & \textbf{0.85} & \textbf{0.02} & \textbf{0.13} & \textbf{0.79} & \textbbf{0.70} & \textbbf{0.45} & \textbbf{0.76} & \textbbf{0.89} & \textbbf{0.60} & \textbbf{0.25} & \textbbf{0.55} & \textbbf{0.53} \\
        \hline
        FAWAC-T & \textbbf{0.24} & \textbbf{0.19} & \textbbf{0.97} & \textbbf{0.13} & \textbbf{0.57} & \textbbf{0.12} & \textbf{0.61} & \textbf{0.89} & \textbf{0.42} & \textbf{0.63} & \textbbf{0.57} & \textbbf{0.79} & \textbbf{0.56} & \textbbf{0.46} \\
        \hline
    \end{tabular}
    }
\end{table}

\vspace{-10pt}
\paragraph{Baselines.} We compare our approach to several safe offline RL baselines that adapt frameworks from safe control, offline RL, and safe RL. FISOR \citep{zheng2024safe} enforces hard constraints to ensure safety under strict cost thresholds but often produces overly conservative policies with limited rewards. CDT \citep{liu2023constrained} uses a decision-transformer framework \citep{chen2021decision} for cost-prompted sequence modeling but struggles to satisfy both reward and cost targets, leading to safety violations under tighter constraints. COptiDICE \citep{lee2022coptidice} and OASIS \citep{yao2024oasis} extend distribution-shaping methods \citep{lee2021optidice} to constrained RL settings. Behavior Cloning (BC), a supervised learning approach, ignores reward and cost labels, often achieving strong rewards but offering no safety guarantees. Finally, CPQ \citep{xu2022constraints} adapts CQL \citep{kumar2020conservative} to constrained RL by penalizing out-of-distribution and unsafe samples but may face trade-offs between safety and performance.

\vspace{-10pt}

\begin{figure}[hbt!]
\centering 
\subfigure[Ball Circle ($\lambda=2.0$)]{\label{fig:AblationCurvesa}\includegraphics[width=75mm]{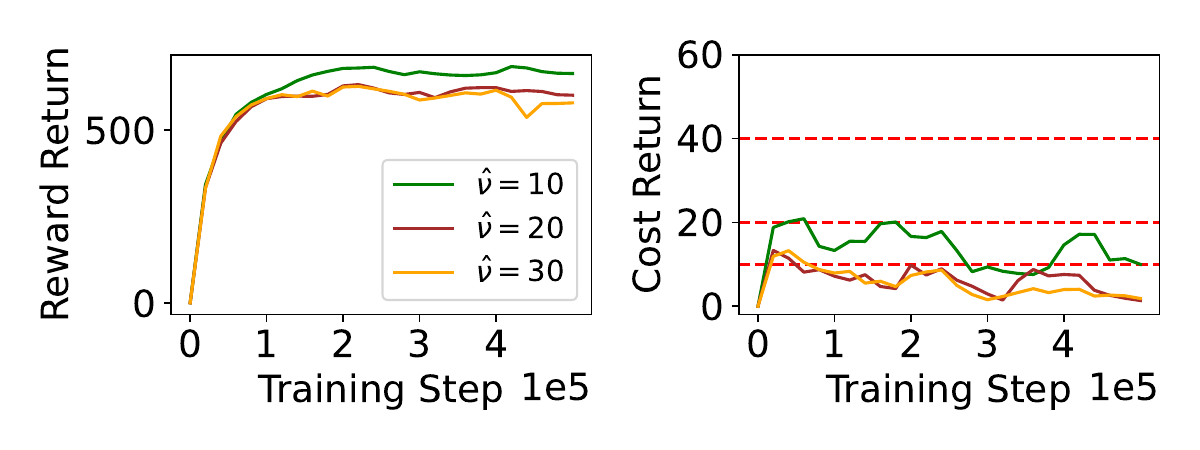}}
\subfigure[Car Circle ($\hat{\nu}=20.0$)]{\label{fig:AblationCurvesb}\includegraphics[width=75mm]{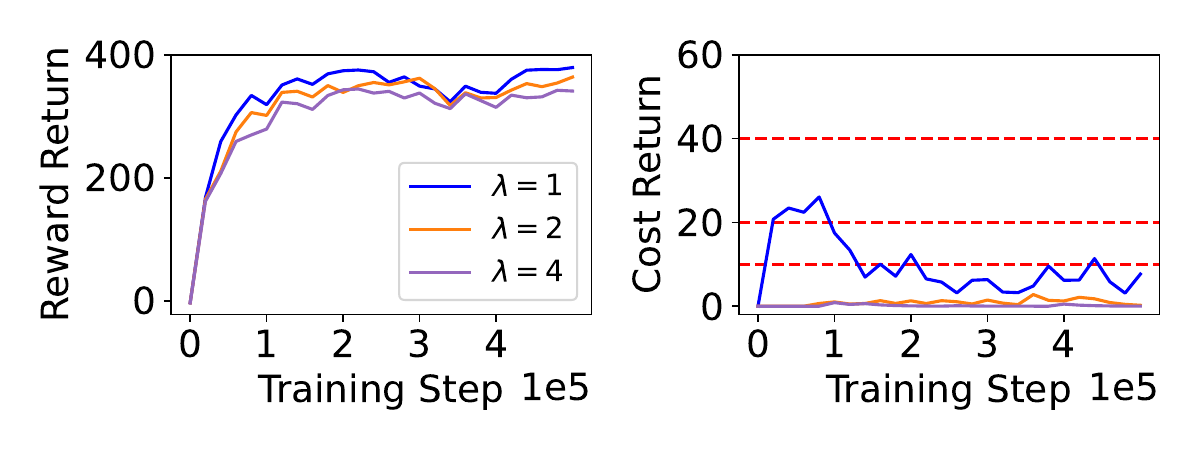}}
\vspace{-7pt}
\caption{Ablation on the FAWAC hyperparameters $\hat{\nu}$ and $\lambda$.}
\label{fig:AblationCurves}
\end{figure}

\subsection{Implementation Details and Ablation}
Our method employs a Lagrange-based formulation, enforcing a soft constraint to ensure that expected safety violation costs remain below predetermined thresholds. However, the discounted cost return threshold ($\kappa$) used in our formulation cannot be directly related to the undiscounted threshold ($\kappa'$) used in empirical evaluations. Following \cite{zhang2023evaluating}, this relationship can be approximated in terms of \textit{planning steps}. In our experiments, we set $\kappa=7.32$, corresponding to planning steps of approximately 30, 100, and 140 for $\kappa'$ values of 10, 20, and 30, respectively. Other key hyperparameters include the temperature $\lambda$ in AWR, the maximum value of $\nu(s)$ in FAWAC-M, and the penalty factor $\hat{\nu}$ in FAWAC-P. The effects of these hyperparameters are shown in Figure \ref{fig:AblationCurves}. Both lower $\hat{\nu}$ and lower $\lambda$ tend to slightly improve the reward-wise performance but can lead to higher cost returns.  In our experiments, we use $\lambda=2$, with $\nu_{max}$ and $\hat{\nu}$ both set to 20.

\section{Conclusion}
We introduced the Feasibility Informed Advantage Weighted Actor-Critic (FAWAC) framework for safe offline reinforcement learning, which incorporates feasibility conditions into the advantage-weighting mechanism. FAWAC, with its variants FAWAC-M and FAWAC-P, effectively balances safety and performance, while FAWAC-T addresses challenges in \textit{tempting} datasets containing high-reward but unsafe trajectories. Demonstrating strong results across diverse robot control tasks, FAWAC provides a scalable approach for safety-critical applications. However, some limitations on the applicability of this work might stem from its reliance on simplifying assumptions about the feasibility of states in the dataset. Addressing these limitations in future work could involve integrating more accurate feasibility estimation techniques, developing adaptive constraint thresholds for dynamic safety requirements, and improving its robustness in online fine-tuning.

\clearpage


\bibliography{bib}

\iftrue
\clearpage
\section{Appendix} \label{appendix_section}
In this section, we present missing analysis and/or proof in the main text, as well as the additional results. In the following, we begin by discussing the approximation of the optimization problem.

\subsection*{Optimization problem approximation}
In this subsection, we present the technical detail about how the original optimization problem, as described in Eq.~\ref{feasible-infeasible-optim} to~\ref{feasible-infeasible-optim-ineq-2}, is approximated by the formulation in Eq.~\ref{achiam-adv-optim-problem} to~\ref{achiam-adv-optim-problem-probab-constraint}. To this end, we define $J(\pi) = \mathbb{E}_{\tau\sim\pi}[R(\tau)]$, which is the expected return. Likewise, we can obtain the cost return $J_c(\pi)=\mathbb{E}_{\tau\sim\pi}[C(\tau)]$. Recalling the definition of reward value function $V^\pi_r(s)=\mathbb{E}_{\tau\sim\pi}[\sum_{t=0}^T\gamma^tr(s_t,a_t)|s_t=s]$ and cost value function $V^\pi_c(s)=\mathbb{E}_{\tau\sim\pi}[\sum_{t=0}^T\gamma^tc(s_t,a_t)|s_t=s]$, we can acquire the following relationships
\begin{equation}\label{eq_31}
    J(\pi) = \mathbb{E}_{s\sim\mathcal{D}}[V^\pi_r(s)],\;J_c(\pi) = \mathbb{E}_{s\sim\mathcal{D}}[V^\pi_c(s)].
\end{equation}
$s\sim\mathcal{D}$ is due to the offline static dataset, instead of the online interactions with the environment. This intuitively makes sense as the expectation of the state value given all states in the offline dataset essentially approximates the expected total return. With this in hand, we recall the original problem formulation in the following
\begin{align*}
\max_{\pi} & \quad \mathbb{E}_{s \sim \mathcal{D}}\left[V_r^\pi(s) \cdot \mathbbm{1}_{s \in \mathcal{S}_f} - V_c^\pi(s) \cdot \mathbbm{1}_{s \notin \mathcal{S}_f}\right]  \\
\text{subject to} & \quad V_c^\pi(s) \leq \kappa, \quad \forall s \in \mathcal{S}_{f0} \\
& \quad D_{\text{KL}}(\pi \parallel \pi_\beta) \leq \delta
\end{align*}
With the assumption that $\mathcal{S}_{f0}=\mathcal{S}_0$ and Eq.~\ref{eq_31}, one can immediately rewrite the above problem as follows
\begin{equation}\label{eq_32}
\begin{split}
    \max_{\pi} & \quad J(\pi)\\
    \text{subject to} & \quad V_c^\pi(s) \leq \kappa, \quad \forall s \in \mathcal{S}_{f0} \\
    & \quad D_{\text{KL}}(\pi \parallel \pi_\beta) \leq \delta
\end{split}
\end{equation}
We still keep the cost constraint in the form of state value, instead of the cost return. The new problem essentially resembles the constrained policy optimization problem in Eq.3 in~\citep{achiam2017constrained}. Since for any two policies $\pi',$ and $\pi$, we have the relationship: $J(\pi')-J(\pi)=\frac{1}{1-\gamma}\mathbb{E}_{s\sim d^{\pi'},a\sim\pi'}[A^\pi(s,a)]$. This will assists in the conversion of the optimization from $J(\pi)$ to $\mathbb{E}_{s\sim d^{\pi'},a\sim\pi'}[A^\pi(s,a)]$. Based on policy performance bounds presented in~\citep{achiam2017constrained}, the maximization on $J(\pi)$ can be transformed into the maximization of the expectation of improvement between two consecutive steps between $\pi^k$ and $\pi^{k+1}$, i.e., $\mathbb{E}_{s\sim d^{\pi^k},a\sim\pi^{k+1}}[A^{\pi^k}(s,a)]$. This is also achieved by the \textit{trust region} method presented in~\citep{schulman2015trust}. Originally, the trust region was proposed to stabilize the policy learning and ensure a feasible policy. However, in our study, we extend this to a \textit{safe trust region} driven by the cost constraint, as shown in $\Pi_f$, resembling a similar formulation in Eq.10 in~\citep{achiam2017constrained}. It should be noted that the significant difference between our setup and that in~\citep{achiam2017constrained} is that our learning setting is no longer online, but offline, which brings about the data distribution drift during the policy learning. As the unknown behavior policy $\pi_\beta$ changed the behavior regularization from $D_{KL}(\pi_{k+1}||\pi^k)$ to $D_{KL}(\pi^{k+1}||\pi_\beta)$. By corollaries 1, 2, and 3 in~\citep{achiam2017constrained}, and the trust region method, we can now rewrite the problem formulation in Eq.~\ref{eq_32} into Eqs.~\ref{achiam-adv-optim-problem} to~\ref{achiam-adv-optim-problem-probab-constraint}.
\subsection*{Proof of Lemma~\ref{lemma_1}}
In this subsection, we provide the proof for Lemma~\ref{lemma_1}.
\begin{proof}
    First, the objective function is linear with respect to $\pi$. As $V^{\pi^k}_c(s)$ is a constant w.r.t $\pi$, Eq.~\ref{eq_14} is linear. The KL divergence in Eq.~\ref{eq_15} is convex w.r.t. $\pi$. Since $\pi$ itself is linear, which is considered convex. Hence, the integral of a convex function in Eq.~\ref{achiam-adv-optim-problem-probab-constraint} is also convex. Thus, the desirable result in obtained. 
\end{proof}
\subsection*{Proof of Theorem~\ref{theorem_2}}
In this subsection, we provide the proof for Theorem~\ref{theorem_2}. 
\begin{proof}
Rewrite $\mathcal{L}$ in the following form:
\begin{equation}
\begin{split}
    \mathcal{L}(\pi,\nu,\lambda,\alpha)=& \nu\bar{\kappa}+\lambda\delta+\alpha+\\&\mathbb{E}_{a\sim\pi(\cdot|s)}[A^{\pi^k}(s, a)-\nu A_c^{\pi^k}(s, a)-\lambda(\text{log}\pi(a|s)-\text{log}\pi_\beta(a|s))]-\alpha\sum_a\pi(a|s)
\end{split}
\end{equation}
We can further rewrite $\mathcal{L}$ to remove constants w.r.t. $\pi(a|s)$ such that
\begin{equation}
    \mathcal{L}(\pi,\nu,\lambda,\alpha)=\sum_a\pi(a|s)[A^{\pi^k}(s, a)-\nu A_c^{\pi^k}(s, a)-\lambda(\text{log}\pi(a|s)-\text{log}\pi_\beta(a|s))-\alpha]
\end{equation}
The last equality holds since the fraction in the expectation is a constant w.r.t $\pi(a|s)$.
    Differentiating $\mathcal{L}(\pi, \nu, \lambda, \alpha)$ w.r.t. $\pi(a|s)$ for some $a$: 
\begin{align}
\frac{\partial \mathcal{L}}{\partial \pi} = A^{\pi^k}(s, a) - \nu A_c^{\pi^k}(s, a) + \lambda \log \pi_\beta(a|s) - \lambda \log \pi(a|s) - \lambda - \alpha.
\end{align}
Setting $\frac{\partial \mathcal{L}}{\partial \pi}$ to zero and rearranging to solve for $\pi(a|s)$ gives:
\begin{equation}
    \pi(a|s)=\pi_\beta(a|s)\text{exp}\bigg(\frac{1}{\lambda} \left( A^{\pi^k}(s, a) - \nu A_c^{\pi^k}(s, a) \right)\bigg)\text{exp}\bigg(-1-\frac{\alpha}{\lambda}\bigg).
\end{equation}
Since $\int_a\pi(a|s)da=1$, the second exponential term is the partition function $Z(s)$ that normalizes the conditional action distribution,
\begin{equation}
    Z(s)=\text{exp}\bigg(1+\frac{\alpha}{\lambda}\bigg)=\int_{a'}\pi_\beta(a'|s)\text{exp}\bigg(\frac{1}{\lambda} \left( A^{\pi^k}(s, a) - \nu A_c^{\pi^k}(s, a) \right)\bigg)da'.
\end{equation}
Hence, the optimal policy $\pi^*(a|s)$ takes the following form
\begin{align}
\pi^*(a|s) = \frac{1}{Z(s)} \pi_\beta(a|s) \exp\left(\frac{1}{\lambda} \left( A^{\pi^k}(s, a) - \nu A_c^{\pi^k}(s, a) \right) \right).
\end{align}
This completes the proof.
\end{proof}
\subsection*{Proof of Proposition~\ref{proposition_3}}
In this subsection, we present the proof for Proposition~\ref{proposition_3}. 
\begin{proof}
    Based on the Corollary 2 from~\citep{achiam2017constrained},we have the following relationship
    \begin{equation}\label{eq_39}
        J_c(\pi')-J_c(\pi)\leq \frac{1}{1-\gamma}\mathbb{E}_{s\sim d^{\pi}(s),a\sim \pi'}\bigg[A^\pi_c(s,a)+\frac{2\gamma \epsilon^{\pi'}_c}{1-\gamma}D_{TV}(\pi'||\pi)(s)\bigg],
    \end{equation}
    where $\epsilon^{\pi'}_c=\max_s|\mathbb{E}_{a\sim\pi'}[A^\pi_c(s,a)]|$, $D_{TV}(\cdot||\cdot)$ signifies the total variation distance between two distributions.
    $\pi^k$ is a feasible point of the objective function in Eq.~\ref{achiam-adv-optim-problem} with objective value 0, such that $\mathbb{E}_{s\sim d^{\pi^k}(s),a\sim\pi^*}[A^{\pi^k}_c(s,a)]\leq 0$. Using the Triangle inequality leads us to 
    \begin{equation}
        D_{TV}(\pi'||\pi)(s)\leq D_{TV}(\pi'||\pi_\beta)(s) + D_{TV}(\pi_\beta||\pi)(s)
    \end{equation}
    Based on the Pinsker's inequality and Jensen's inequality, we have the following relationships
    \begin{equation}
        \mathbb{E}_{s\sim d^{\pi_\beta}(s)}[D_{TV}(\pi'||\pi_\beta)(s)]\leq \mathbb{E}_{s\sim d^{\pi_\beta}(s)}\bigg[\sqrt{\frac{1}{2}D_{KL}(\pi'||\pi_\beta)(s)}\bigg]\leq \sqrt{\frac{1}{2}\mathbb{E}_{s\sim d^{\pi_\beta}(s)}[D_{KL}(\pi'||\pi_\beta)(s)]},
    \end{equation}
    and 
        \begin{equation}
        \mathbb{E}_{s\sim d^{\pi_\beta}(s)}[D_{TV}(\pi_\beta||\pi)(s)]\leq \mathbb{E}_{s\sim d^{\pi_\beta}(s)}\bigg[\sqrt{\frac{1}{2}D_{KL}(\pi_\beta||\pi)(s)}\bigg]\leq \sqrt{\frac{1}{2}\mathbb{E}_{s\sim d^{\pi_\beta}(s)}D_{KL}(\pi_\beta||\pi)(s)},
    \end{equation}
    which implies that 
$\mathbb{E}_{s\sim d^{\pi_\beta}(s)}[D_{TV}(\pi'||\pi_\beta)(s)]\leq \sqrt{\frac{1}{2}\mathbb{E}_{s\sim d^{\pi_\beta}(s)}[D_{KL}(\pi'||\pi)(s)]}\leq \sqrt{\frac{\delta}{2}}$. Similarly, $\pi^k$ is feasible point such that $\mathbb{E}_{s\sim d^{\pi_\beta}(s)}[D_{TV}(\pi_\beta||\pi^k)(s)]\leq \sqrt{\frac{1}{2}\mathbb{E}_{s\sim d^{\pi_\beta}(s)}[D_{KL}(\pi_\beta||\pi^k)(s)]}\leq \sqrt{\frac{\delta}{2}}$. Also,
it should satisfy $J_c(\pi^k)\leq \kappa$. Setting $\pi'=\pi^*$, using Eq.~\ref{eq_31}, and substituting into Eq.~\ref{eq_39} yields the desirable result.
\end{proof}

\fi

\end{document}